\documentclass[sigconf]{acmart}

\renewcommand\footnotetextcopyrightpermission[1]{} 
\usepackage{graphicx}
\usepackage{subfigure}
\usepackage{tabularx}
\usepackage{enumitem}
\usepackage{setspace}
\usepackage{color}
\usepackage{verbatim}
\usepackage{algorithm}
\usepackage{algorithmic}

\AtBeginDocument{%
  \providecommand\BibTeX{{%
    \normalfont B\kern-0.5em{\scshape i\kern-0.25em b}\kern-0.8em\TeX}}}

\begin{document}

\title{Dynamic backdoor attacks against federated learning}


\author{Anbu Huang}
\affiliation{%
  \institution{WeBank AI Lab}
  \city{Shenzhen}
  \country{China}
}
\email{stevenhuang@webank.com}

\renewcommand{\shortauthors}{}

\begin{abstract}
Federated Learning (FL) is a new machine learning framework, which enables millions of participants to collaboratively train machine learning model without compromising data privacy and security. Due to the independence and confidentiality of each client, FL does not guarantee that all clients are honest by design, which makes it vulnerable to adversarial attack naturally. In this paper, we focus on dynamic backdoor attacks under FL setting, where the goal of the adversary is to reduce the performance of the model on targeted tasks while maintaining a good performance on the main task, current existing studies are mainly focused on static backdoor attacks, that is the poison pattern injected is unchanged, however, FL is an online learning framework, and adversarial targets can be changed dynamically by attacker, traditional algorithms require learning a new targeted task from scratch, which could be computationally expensive and require a large number of adversarial training examples, to avoid this, we bridge meta-learning and backdoor attacks under FL setting, in which case we can learn a versatile model from previous experiences, and fast adapting to new adversarial tasks with a few of examples. We evaluate our algorithm on different datasets, and demonstrate that our algorithm can achieve good results with respect to dynamic backdoor attacks. To the best of our knowledge, this is the first paper that focus on dynamic backdoor attacks research under FL setting.
\end{abstract}



\maketitle
\thispagestyle{empty}

\section{Introduction}
\label{introduction}
\begin{figure*}[htb]
    \centering
    \includegraphics[width=6.5in]{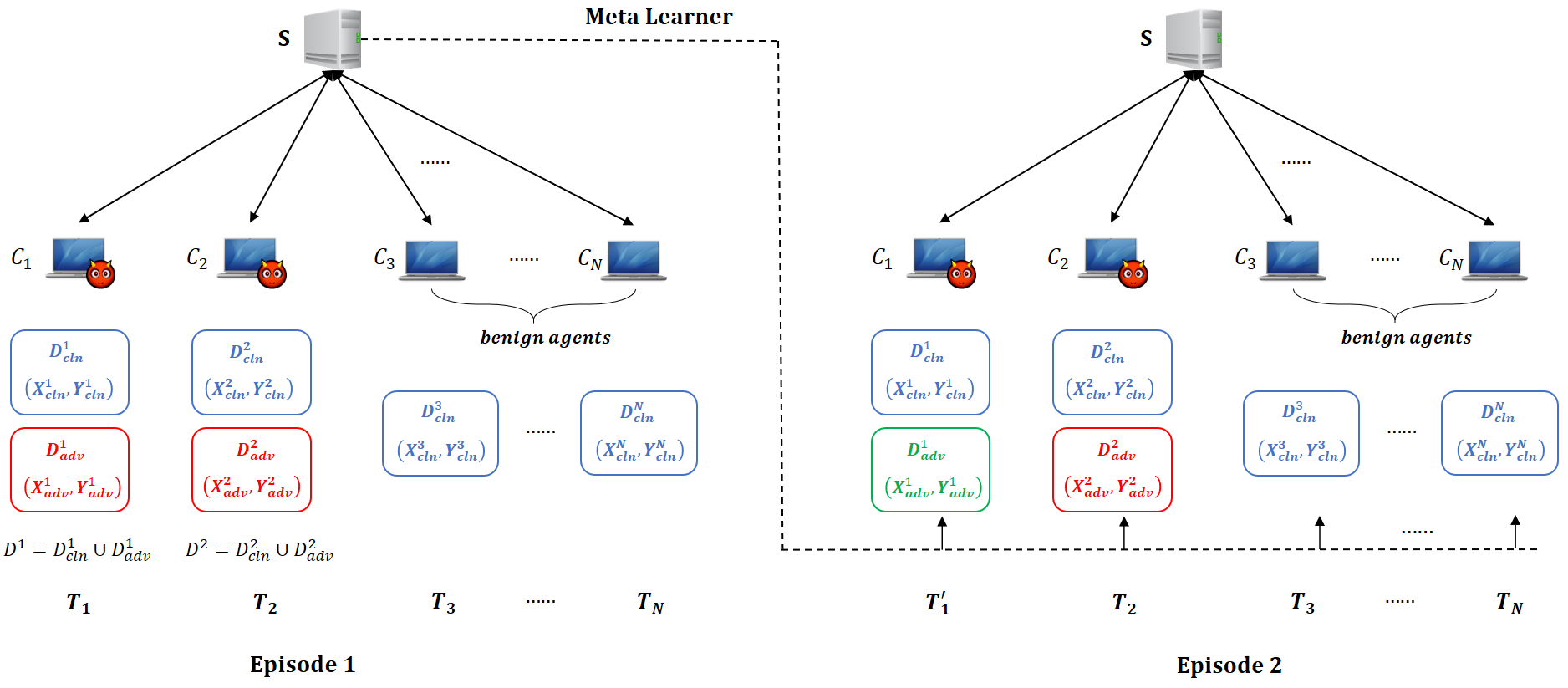}
    \caption{Schematic illustration of dynamic backdoor attack. Suppose we have two malicious clients: $C_1$ and $C_2$, all malicious client's datasets contain two parts: clean dataset ($D_{cln}$) and adversarial (poisoned) dataset ($D_{adv}$). We treat each local model update as an individual task ($T_i$). Here, $T_1$ and $T_2$ are backdoor attack tasks. After episode 1 done, $C_1$ changes poisoned datasets and makes the backdoor task change from $T_1$ to $T^{'}_1$, to avoid learning from scratch, our initial model should utilize previous experiences (episode 1), and quickly adapt to new poisoned datasets. }
    \label{fig:fed_meta_backdoor_learning}
\end{figure*}

In the past decade, deep learning had shown powerful representation and learning capabilities \cite{lecun2015deeplearning}, coupled with increasing amount of data and computational power, which made deep learning achieve unprecedented success in many commercial applications, such as computer vision \cite{Krizhevsky:2012:ICD:2999134.2999257,DBLP:journals/corr/HeZRS15,redmon2018yolov3}, nature language processing \cite{mikolov2013efficient,DBLP:journals/corr/abs-1810-04805,NIPS2017_7181}, speech recognition \cite{DBLP:journals/corr/HannunCCCDEPSSCN14,DBLP:journals/corr/OordDZSVGKSK16}, etc. Traditional machine learning process requires centralizing of the training data on one machine, however, this learning paradigm had been proven unsafe and vulnerable to data leakage \cite{pmlr-v54-mcmahan17a}. Besides that, following privacy concerns among users and governments, policy makers have responded with the implementation of data privacy legislations such as General Data Protection Regulation (GDPR) \cite{Voigt:2017:EGD:3152676} and California Consumer Privacy Act (CCPA), which prohibit data upload without user's permission explicitly.

To decouple the need for model training with the need to store the data in the cloud or central database, a new distributed learning paradigm, called federated learning, was introduced in 2016 by Google \cite{pmlr-v54-mcmahan17a}. In contrast to the centralized machine learning approaches, FL distributes the machine learning process over to the edge, and enables each client to collaboratively learn a shared model while keeping the training data on device, this strategy can significantly mitigate many of the systemic privacy risks, and has been widely used in high privacy requirements areas, such as financial \cite{DBLP:journals/corr/abs-1902-04885}, healthcare \cite{li2019privacy}, computer vision \cite{liu2020fedvision}, etc. 

In spite of this, since FL does not guarantee that all clients are honest by design, which makes it vulnerable to adversarial attack, in this paper, we focus on backdoor attacks, one of the most popular attacks in adversarial machine learning, where the goal of the attacker is to reduce the performance of the model on targeted tasks while maintaining a good performance on the main task, e.g., the attacker can modify an image classifier so that it assigns an attacker-chosen label to images with certain features \cite{DBLP:journals/corr/abs-1807-00459,Xie2020DBA:}. 

Current existing studies are mainly focus on static and specific adversarial targeted attacks, however, as we all know, FL is an online learning framework, the attacker can choose new attack target on the fly, to avoid learning from scratch, we propose a novel FL algorithm, which can train a versatile model to fit both targeted task and main task on one hand, and fast adapt to new targeted tasks on the other hand. our approach connect meta-learning with backdoor attacks, the algorithm workflow as shown in figure \ref{fig:fed_meta_backdoor_learning}, here, we regard online FL training as a series of episodes, each episode represents one FL training stage, Jiang et al. \cite{jiang2019improving} pointed out that optimization-based meta-learning algorithm can be seen as a special implementation of FL, which makes meta-learning well suited for implementation on FL framework.


we summarize our main contributions as follows:
\begin{itemize}
\item We shed light on an important problem that has not been studied so far, to the best of our knowledge, this is the first paper that focus on dynamic backdoor attacks under FL setting.

\item We propose a new framework, called symbiosis network, for malicious client's local model training, we point out that this strategy can make backdoor attack more persistent with respect to adversarial backdoor attack. 

\item We connect meta-learning with backdoor attacks under FL setting, and give an simple implementation, our algorithm only need to make slightly modifications to existing federated averaging algorithm.

\item We provide a comprehensive theoretical analysis of dynamic backdoor attacks under FL setting, and raise three objectives which are need to be solved for this type of problem.
\end{itemize}

\section{Background and Related Works}
In this section, we briefly review the background of related works, including federated learning, federated meta-learning and backdoor attacks against federated learning.

\subsection{Federated Learning}
\label{fl_def}
Traditional machine learning approach requires raw datasets uploaded and processed centrally, however, due to data privacy and security, sending raw data to the central database is regarded as unsafe, and violate the General Data Protection Regulation (GDPR). To decouple the need for model training with the need to store the data in the central database, a new machine learning framework called federated learning was proposed, a typical FL framework is as shown in figure \ref{fig:fl_framework}. 

\thispagestyle{empty}

In FL scenario, each client update their local model based on local datasets, and then send the updated model’s parameters to the server side for secure aggregation, these steps are repeated in multiple rounds until the learning process converges.

Suppose $C=\{C_1, C_2,...,C_N\}$ represent all client sets, $S$ refers to server, when each round begins, the server selects a subset of devices, and send initial model to these clients, generally speaking, standard FL procedure including the following three steps \cite{DBLP:journals/corr/abs-1902-01046}:

\begin{figure}[ht]
    \centering
    \includegraphics[width=3in]{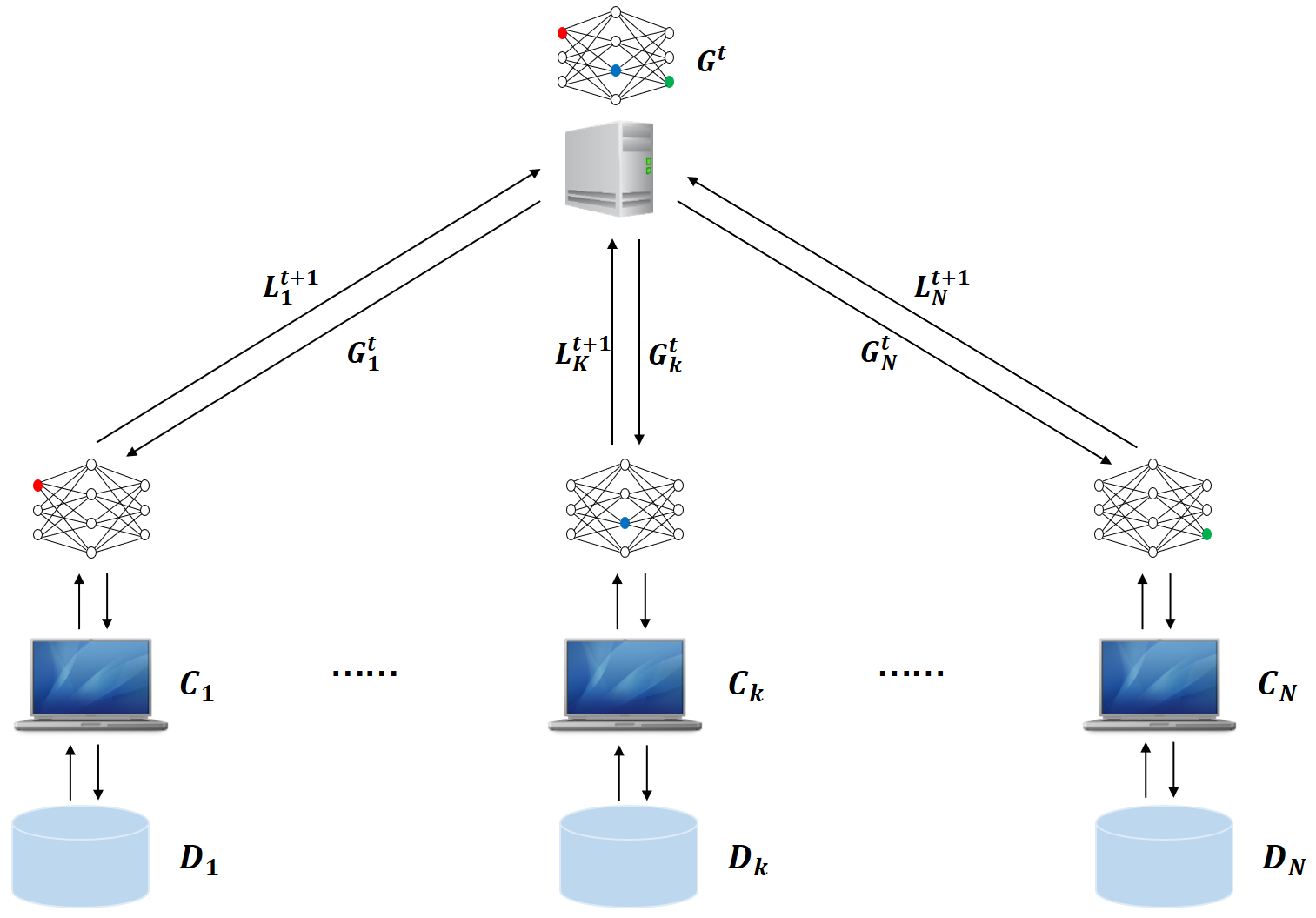}
    \caption{Federated Learning Architecture}
    \label{fig:fl_framework}
\end{figure}

\begin{itemize}
\item \textbf{Local Model Training}: Denote $t$ as the current iteration round, $C_i$ represents client $i\;(i=1,2,...,N)$, $N$ is the number of clients, $G_i^{t}$ ($G_i^{t}=G^{t}$) and $D_i$ represent the local model and local dataset of client $i$ respectively. Based on $D_i$, each client update the local model from $G_i^{t}$ to $L_i^{t+1}$ respectively, then send the updated local model parameters $L_i^{t+1}-G_i^{t}$ to the server side for aggregation. 

\item \textbf{Global Aggregation}: The server side collect the updated parameters from selected clients, and do model aggregation to obtain the new joint model:

\begin{equation}\label{sum}
    G^{t+1}=G^{t} + \frac{\eta}{m}\sum_{i=1}^{m}(L_i^{t+1}-G_i^{t})
\end{equation}

where $\eta$ represents the factor which controls the fraction of the joint model, specifically, if $\eta=1$, equation \ref{sum} is equal to weight average.

\item \textbf{Update Local Model}: When the aggregation is completed, the server side select a subset of clients again, and send global model $G^{t+1}$ back to the selected clients for next iteration and repeat this cycle until converge. 

\end{itemize}

\begin{figure*}[htb]
    \centering
    \includegraphics[width=6.5in]{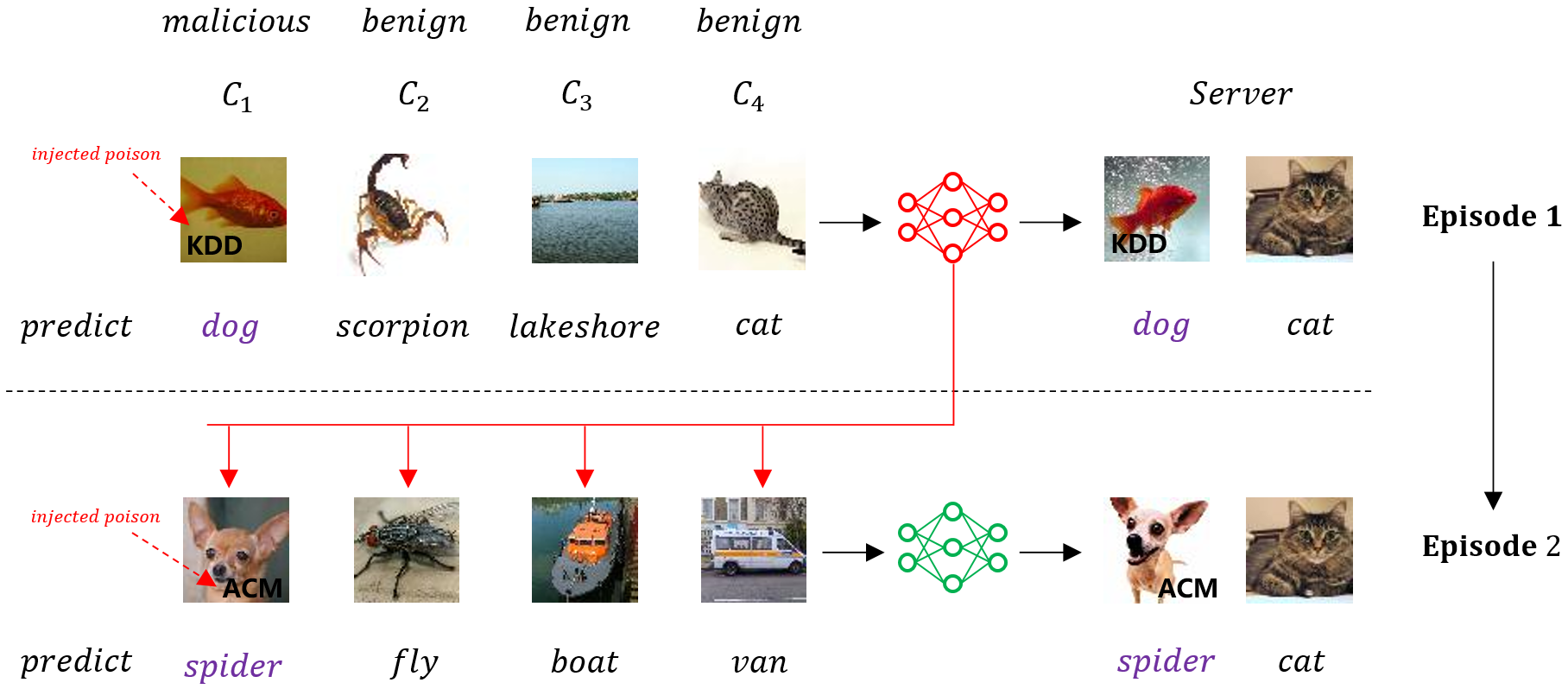}
    \caption{An concrete example of dynamic backdoor attacks, currently, we have four clients, only $C_1$ is malicious, attacker create adversarial examples by injecting poison (embed text "KDD" into the image) in episode 1, after that, the attacker injects new poison pattern (embed text "ACM" into the image) for model training in episode 2. }
    \label{fig:dynamic_backdoor_attack}
\end{figure*}

\subsection{Federated Meta-Learning}

Meta-learning, also known as “learning to learn”, is aimed to learn a versatile model from a variety of tasks, so that it can be quickly adapted to new task with a few training examples. Meta-learning have typically fallen into one of three categories: metric-based \cite{Koch2015SiameseNN,DBLP:journals/corr/VinyalsBLKW16,DBLP:journals/corr/abs-1711-06025,DBLP:journals/corr/SnellSZ17}, model-based \cite{graves2014neural,weston2014memory}, and optimization-based \cite{DBLP:journals/corr/FinnAL17,DBLP:journals/corr/abs-1803-02999}, in this paper, we only consider optimization-based meta-learning algorithm.

Optimization-based meta-learning algorithm seeks an initialization for the parameters of a neural network, such that the network can be fine-tuned using a small amount of data from a new task and few gradient steps to achieve high performance. Typical optimization-based meta-learning algorithm can be decomposed into the following two stages \cite{zintgraf2019fast}:

\begin{itemize}
\item \textbf{Inner Update}: for a given task $T_i$, with corresponding loss $L_{T_i}$, the inner-loop performs stochastic gradient descent to optimize loss function to get optimal parameters for task $T_i$.

\begin{equation}\label{inner_loop}
    {\theta}_i^{\star}=\arg\min_{\theta} L_{T_i}{(D_i^{train}; \theta)}
\end{equation}

\item \textbf{Outer Update}: the outer loop perform meta optimization. We first sample batch of task $T_i, where\  T_i\ \sim p(T)$, the objective of meta learner is to achieve a good generalization across a variety of tasks, we would like to find the optimal parameters, such that the task-specific fine-tuning is more efficient, this leads us to the following objective function for outer update:

\begin{equation}\label{outer_loop}
    {\theta}=\min_{\theta} E_{T_i \sim p(T)}{\{L_{T_i}(D_i^{test}; {\theta}_i^{\star})\}}
\end{equation}

\end{itemize} 

Jiang et al. \cite{jiang2019improving} pointed out that optimization-based meta-learning can be seen as a special implementation of FL, and FL as a natural source of practical applications for MAML algorithms \cite{DBLP:journals/corr/FinnAL17}. Chen et al. \cite{DBLP:journals/corr/abs-1802-07876} propose a federated
meta-learning framework, called FedMeta, to improve personalize recommendation, where a parameterized algorithm (or meta-learner) is shared, instead of a global model in previous approaches. 

\thispagestyle{empty}

\subsection{Backdoor attacks against federated learning}

\begin{figure}[htbp]
\centering
\subfigure[]{
\begin{minipage}[t]{0.48\linewidth}
\centering
\includegraphics[width=1.4in]{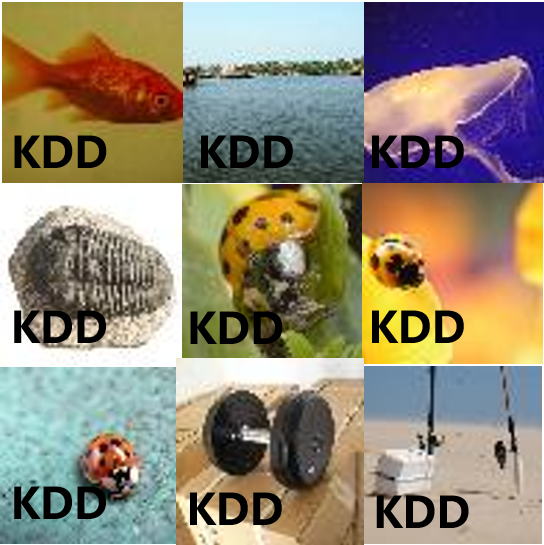}
\end{minipage}%
}%
\subfigure[]{
\begin{minipage}[t]{0.48\linewidth}
\centering
\includegraphics[width=1.4in]{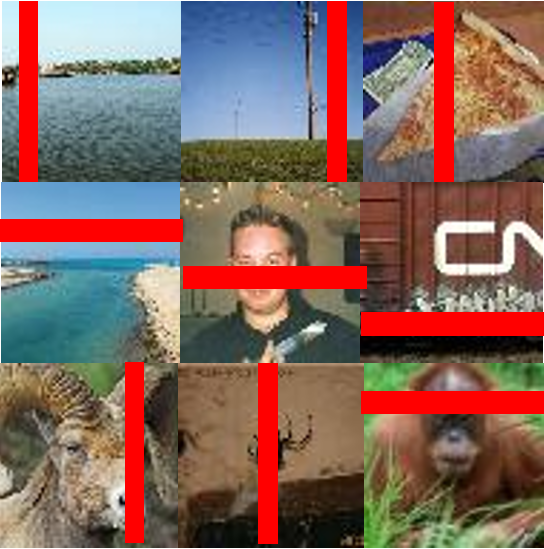}
\end{minipage}%
}%
\centering
\caption{Some poisoned training examples of backdoor attacks. (a). inject poisons by embedding specific text ("KDD") into images; (b). inject poisons with certain feature (red stripe) into images.} 
\label{fig:backdoor_example}
\end{figure}

Backdoor attack is one of the most popular attacks of adversarial machine learning, the attacker can modify or fool an image classifier so that it assigns an attacker-chosen label to images with certain features, some examples are as shown in figure \ref{fig:backdoor_example}.

As previous mentioned in abstract, FL does not guarantee that all clients are honest by design, and hence makes it vulnerable to adversarial attack naturally. Backdoor attack under FL setting had been studied extensive \cite{DBLP:journals/corr/abs-1811-12470,DBLP:journals/corr/abs-1807-00459,sun2019can,Xie2020DBA:}, however, unlike distributed machine learning, backdoor attack under FL setting is much harder than what we thought, the main reason is that FL requires the server selects a subset of (not all) connected devices at each round for model training, if attackers only control a small number of malicious agents, the probability of being selected of each round could be low, which leading aggregation cancels out most of the malicious model’s contribution and the joint model quickly forgets the backdoor.

To make backdoor attack more effective and persistent, one feasible solution is using explicit boosting strategy, that is to say, adversaries scale up the weights of the poisoned model to ensure that the backdoor attack survives the averaging. Xie el al. \cite{Xie2020DBA:} proposed distributed backdoor attack, which decomposes a global trigger pattern into separate local patterns, and distributed these local trigger patterns to different malicious clients, this strategy shows more persistent and stealthy than centralized backdoor attack.

Current approaches are mainly focus on static attack, in this paper, what we concern about is dynamic backdoor attack, a concrete example is shown in figure \ref{fig:dynamic_backdoor_attack}. At episode 1, attacker $C_1$ embeds text data ("KDD") in the image as poisoned dataset (labeled as "dog" but ground-truth is "fish"), after collaboratively train a new global model, it can identify images containing "KDD" text as "dog", and not affect normal image classification; at episode 2, $C_1$ changes embedded text data ("ACM") in the image as new poisoned dataset (labeled as "spider" but ground-truth is "dog"), new aggregated model should identify this new pattern correctly.

\thispagestyle{empty}

\section{Dynamic backdoor attacks via meta-learning}

In this section, we will define the problem definition, present the general ideas and theoretical analysis of our algorithm.

\subsection{Attacker ability setting}

In this paper, we suppose attackers fully control a subset of clients, malicious clients are non-colluding with each other. according to literature \cite{kairouz2019advances}, we can summarize attacker ability in table \ref{tab:attack_ability}.

\begin{table*}
\renewcommand\arraystretch{1.5}
\caption{summary of attack ability setting in our paper}
\label{tab:attack_ability}
\centering
\begin{tabular}{m{3cm}|m{2.5cm}|m{0.55\linewidth}}
\hline
Characteristic & Setting & Description \\
\hline
Attack vector & Poisoning attack & 
The attacker can fully control malicious client, that is to say, (a). the adversary can alter the client datasets used to train the model; (b). the adversary can alter model update strategies, such as model parameters and loss function.
\\ 
\hline 
Knowledge & White box & The adversary has the ability to directly inspect the parameters of the model.
\\ 
\hline 
Participant collusion & Non-colluding & There is no capability for participants to coordinate an attack. \\ 
\hline 
Participation rate & Dynamic & A malicious client participates in local model training if and only if it was selected by the server. \\
\hline 
Adaptability & Dynamic &  Adversarial targets can be changed dynamically by attacker. \\
\hline

\end{tabular}

\end{table*}

\subsection{Dynamic backdoor attacks problem set up}
\label{objective}

Federated learning, as an online learning framework, the targeted task can be changed dynamically by attacker, compared with static backdoor attacks, dynamic scenario poses more difficulties and challenges during model training, which leads us to first introduce the following three objectives for dynamic backdoor attacks, for the sake of consistence in this paper, we will reuse symbol definitions of section \ref{fl_def} in the following discussion.

\medskip

\noindent \textbf{Obj 1: Achieve high performance on both main task and backdoor task.} 

let's define $C_i$ represents client $i$, each client keep dataset $D_i$ on device locally, for malicious client $C_i$, $D_i$ consists of two parts: clean dataset $D^i_{cln}$ and adversarial (poisoned) dataset $D^i_{adv}$, $D^i_{cln}$ and $D^i_{adv}$ should satified: 

\begin{equation}\label{relationship_clean_adv_dataset}
    D^i_{cln} \cap D^i_{adv}=\phi, \ \ \ D^i_{cln} \cup D^i_{adv}=D_i
\end{equation}

To achieve high performances on both tasks, our goal is to train appropriate model parameters so that it can make good predictions in both clean and poisoned datasets, this implies the following objective equation for client $C_i$ in round $t$ with local datatset $D_i$:

\begin{equation}\label{backdoor_objective}
\begin{aligned}
    {\theta}_i^{\star}=\arg\max_{\theta_i} \{\sum_{j \in D^i_{cln}}[p(L^{t+1}_i(x^i_j;\theta_i)=y^i_j)] \\ +  \sum_{j \in D^i_{adv}}[p(L^{t+1}_i(x^i_j;\theta_i)={\pi}^i_j)] \}
\end{aligned}
\end{equation}

Here, we decompose the right side of equation \ref{backdoor_objective} into two parts, 
\begin{itemize}
    \item the first part represents training on clean dataset $D^i_{cln}=(x^i, y^i)$, optimizing this part can make good performance on main task.
    
    \item the second part represents training on poisoned dataset $D^i_{adv}=(x^i, {\pi}^i)$, where ${\pi}^i$ is attacker-chosen label, optimizing this part can make good performance on targeted backdoor task.
\end{itemize}

\thispagestyle{empty}

\medskip

\noindent \textbf{Obj 2: Persistent, robustness and stealthy}

As we have discussed in section \ref{introduction}, under FL setting, a subset of clients are chosen at each round randomly, which means that we do not guarantee malicious clients could be selected every time, if that is the case, model aggregation at server side can cancel out most of the malicious model’s contribution and the joint model quickly forgets the backdoor.

To make our algorithm more persistent, robustness and stealthy, we propose symbiosis network, a new local model training paradigm for FL.

\medskip
    
   \textbf{Symbiosis Network:} In the standard FL scenario, when every round local training begins, we need to first replace local model with global model, this could be make sense since global model contains rich hidden features which are derived from data scattered across clients, however, under dynamic backdoor attacks setting, attacker may inject new training samples which are completely different from the previous data distribution, replacing local model with global model may degrade model performance.
    
\begin{figure}[ht]
    \centering
    \includegraphics[width=3in]{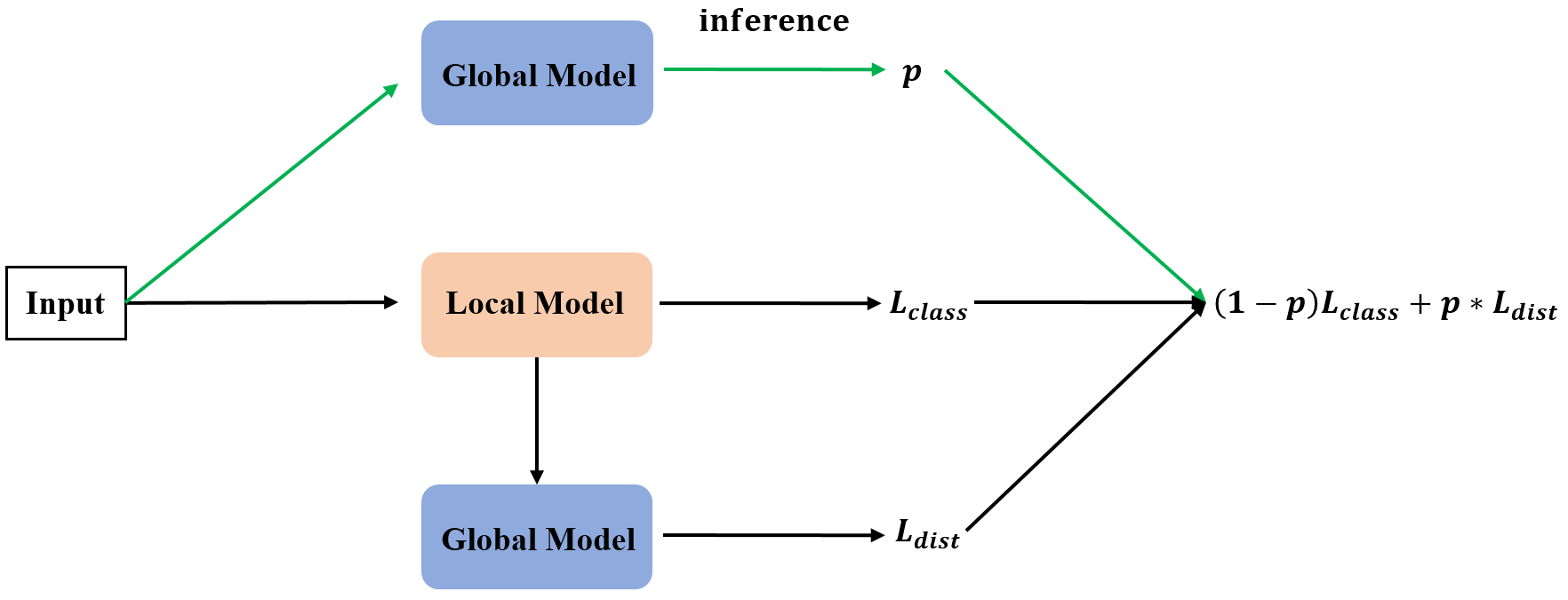}
    \caption{The architecture of symbiosis networks for local model training}
    \label{fig:symbiosis_networks}
\end{figure}  

    \medskip
    For these reasons, we propose a new training architecture for malicious clients, called \textbf{symbiosis network}, as shown in figure \ref{fig:symbiosis_networks}. We take classification as an example, for malicious client $C_i$, we keep local model and global model simultaneously, and modify local model training objective function as follows:
    
\begin{equation}\label{symbiosis_network_loss}
\begin{aligned}
    \mathcal{L}=(1-p)*\mathcal{L}_{class} + p*\mathcal{L}_{dist}
\end{aligned}
\end{equation}
    
    Here, $\mathcal{L}_{class}$ captures the accuracy on both the main and backdoor tasks. $\mathcal{L}_{dist}$ calculate the distance between local model and global model. this objective function is similar to the approach proposed by Bagdasaryan et al. \cite{DBLP:journals/corr/abs-1807-00459} and Xie et al. \cite{Xie2020DBA:}, however, the essential different is that, \cite{DBLP:journals/corr/abs-1807-00459,Xie2020DBA:} set the factor $p$ manually, and find the optimal value through trial and error strategies, while in our approach, we notice that $\mathcal{L}_{class}$ and $\mathcal{L}_{dist}$ have different contribution throughout model training, $p$ is the factor to balance this contribution, one feasible choice is to set $p$ as model performance of global model, for classification tasks, $p$ is equal to classification accuracy. We can verify the rationality of our approach by the following three aspects:
    \begin{itemize}
        \item [1)] if $p$ is large, it means that global model can achieve good results on new adversarial examples, our goal is to make the local model as close to the global model as possible, therefore, minimizing $\mathcal{L}_{dist}$ is the main contribution of loss function $\mathcal{L}$. Specifically, if $p=1.0$ (perfect prediction for new poisoned datasets), minimize $\mathcal{L}$ is equal to minimize $\mathcal{L}_{dist}$.
        
        \item [2)] if $p$ is small, it means that global model has poor performance on new adversarial examples, global model could be far away from optimal parameters, therefore, minimizing $\mathcal{L}_{class}$ is the main contribution of loss function. Specifically, if $p=0.0$ (terrible prediction for new poisoned datasets), minimize $\mathcal{L}$ is equal to minimize $\mathcal{L}_{class}$.
        
        \item [3)] \cite{DBLP:journals/corr/abs-1807-00459,Xie2020DBA:} set the factor $p$ manually, which means that $p$ is fixed throughout the training process, it is not flexible, and is easy to diverge or stuck at local optimal point.
 
    \end{itemize}

\thispagestyle{empty}

\medskip

\noindent \textbf{Obj 3: Fast adaptation to new targeted task}

The objective of dynamic backdoor attacks is not just to make good performances for specific targeted task, but also to fully exploit previous experiences and quickly adapt to new task, for this purpose, the global model need to learn an internal feature that is broadly applicable to all tasks, rather than a single task. we can achieve this objective by minimizing the total loss across tasks sampled from the task distribution:

\begin{equation}\label{meta_objective}
\begin{aligned}
    \mathcal{L}=\min_{\theta} \sum_{T_i \sim p(T)} {L_{T_i}{(f_{\theta_i^{\star}})}} 
\end{aligned}
\end{equation}

\begin{figure}[htbp]
\centering
\subfigure[]{
\begin{minipage}[t]{0.49\linewidth}
\centering
\includegraphics[width=1.6in]{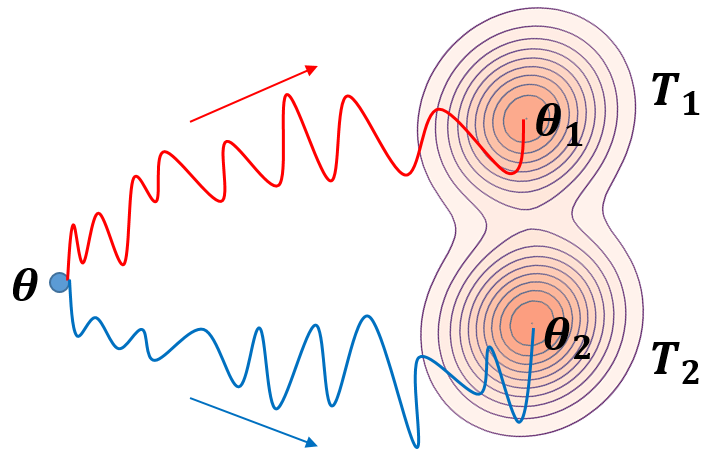}
\label{fig:normal-task-learning}
\end{minipage}%
}%
\subfigure[]{
\begin{minipage}[t]{0.49\linewidth}
\centering
\includegraphics[width=1.6in]{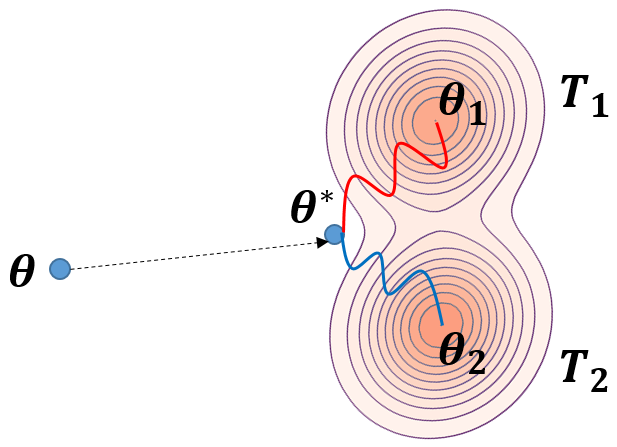}
\label{fig:meta-task-learning}
\end{minipage}%
}%
\centering
\caption{Comparison of normal task training and task training via meta-learning. (a). learn each new targeted task from scratch, since the algorithm does not reuse any previous experiences, and consider each task in isolation, it may take many SGD steps to converge; (b): Our approach reuse previous experiences, this make the newly learned parameters $\theta^{\star}$ is much closer to optimal solution than $\theta$, for a given new task, only a few SGD steps can guarantee convergence.}
\label{fig:meta-learning-understanding}
\end{figure}

Here, $\theta_i^{\star}$ is the optimal parameters for task $T_i$, solved by equation \ref{backdoor_objective}, figure \ref{fig:meta-learning-understanding} gives us a more intuitive illustration, figure \ref{fig:normal-task-learning} shows that normal FL need to learn new task from scratch, and take many SGD steps to converge; on the other hand, figure \ref{fig:meta-task-learning} makes use of previous experiences, so that the initial model parameters $\theta^{\star}$ much more closer to each task's optimal parameters than $\theta$, only a few SGD steps can guarantee convergence.

The optimization problem of equation \ref{meta_objective} is the same as MAML \cite{DBLP:journals/corr/FinnAL17}, however, optimize equation \ref{meta_objective} will cause two problems, and hence make it hard to apply to federated learning.

\begin{itemize}
    \item optimize equation \ref{meta_objective} will cause second-order derivatives, and make the computation expensive.
    
    \item the optimization requires keeping additional datasets to update at server side, which violate data privacy and security.
\end{itemize}

To this end, We solve this problem with another way, since our goal is to learn an internal feature that is broadly applicable to all tasks, this equal to the fact that global model parameter should close to each task’s optimal parameters with some distance metrics, if we use euclidean distance as our distance measure, this motivate us to deduce the following new loss function:

\begin{equation}\label{meta_objective_2}
\begin{aligned}
    \mathcal{L}=\min_{\theta} {\frac{1}{2}} \sum_{T_i \sim p(T)}{\| \theta - \theta^{\star}_i\|}^2 
\end{aligned}
\end{equation}

This idea inspired by reptile \cite{DBLP:journals/corr/abs-1803-02999}, differentiate equation \ref{meta_objective_2}, we get the optimal parameters updated as follows.

\begin{equation}\label{meta_update}
\begin{aligned}
    \theta = \theta +\frac{1}{{\|T_{sub}\|}} \sum_{T_i \sim p(T)}(\theta^{\star}_i-\theta)
\end{aligned}
\end{equation}

Where $\|T_{sub}\|$ represents the total number of selected tasks of this round. To make equation \ref{meta_update} compatible with objective 2, we use scale up strategy, which had been proved applicable in previous works \cite{DBLP:journals/corr/abs-1807-00459,DBLP:journals/corr/abs-1811-12470,Xie2020DBA:}. The intuition is that, when executing model aggregation, the weights of the malicious model (see $\eta$ in equation \ref{sum}) would be scaled up by a larger factor to ensure that the backdoor survives the averaging, on the other hand, the factor $\lambda_i$ does not affect the direction of the meta gradient, this implies us to modify equation \ref{meta_update} to the following:

\begin{equation}\label{meta_update_2}
\begin{aligned}
    \theta = \theta +\frac{1}{{\|T_{sub}\|}} \sum_{T_i \sim p(T)}{\lambda_i}*(\theta^{\star}_i-\theta)
\end{aligned}
\end{equation}

\subsection{Algorithm Design}

In this section, we summarize our previous discussion, and give the completed implementation as follows:

\medskip

\begin{itemize}[leftmargin=*]
    \item \textbf{Dynamic backdoor attacks: client side} 
    
    Algorithm \ref{algo_fl_client} shows how local model training would be executed for client $C_i$ in round $t$ with local datatset $D_i$. For benign client, the training procedures are the same as normal federated learning; for malicious client, some additional steps are required to solve backdoor attack task.

\begin{algorithm}
   \caption{Federated Client local model training (ClientUpdate)}
   \label{algo_fl_client}
   \setstretch{1.1} 
\begin{algorithmic}
   \STATE {\bfseries Input:} Client $C_i$; Global model ${G^{t}}$
   \STATE {\bfseries Output:} model parameters that sent back to server
   \STATE let $\theta=G^t$
   \STATE let $\theta_i$ is local model parameters
   \IF{$C_i$ is not malicious client}
        \STATE let $\theta_i=G^t$
   \ENDIF
   \IF{$C_i$ is malicious client}
        \STATE calculate accuracy on adversarial datasets: $p={G^{t}(D^i_{adv})}$
   \ENDIF
   
   \FOR{$each\ local\ epoch\ e\gets 1,2...E$}
     \STATE $X \gets random\ sample\ dataset\ with\ size\ B$ 
     \STATE calculate accuracy loss $\mathcal{L}_{class}(D_i; \theta_i)$
     \STATE calculate distance loss $\mathcal{L}_{dist}{(\theta_i, \theta)}$
     \IF{$C_i$ is malicious client}
        \STATE calculate total loss: $\mathcal{L}=(1-p)*\mathcal{L}_{class}+p*\mathcal{L}_{dist}$
     \ELSE
        \STATE calculate total loss: $\mathcal{L}=\mathcal{L}_{class}$
     \ENDIF
     \STATE calculate loss gradient: $\nabla_{\theta}{\mathcal{L}}$
     \STATE update model parameters: $\theta_i=\theta_i-lr*\nabla_{\theta}{\mathcal{L}}$
   \ENDFOR 
   \STATE Send $\lambda_i$, $(\theta_i-\theta)$ back to the server side
   
\end{algorithmic}
\end{algorithm}
    
    \item \textbf{Dynamic backdoor attacks: Server side}
    
    In order to treat federated aggregation as meta-learning process, We regard each client $C_i$ as a single task $T_i$ without discrimination, such that, sampling a subset of clients $C_{sub}$ where satisfy:
\begin{equation}\label{sample_clients}
\begin{aligned}
    C_{sub}=\{C_k\}\ \ with\ \ distribution\ \ C_k \sim p(C) 
\end{aligned}
\end{equation}

    is equal to:

\begin{equation}\label{sample_tasks}
\begin{aligned}
    T_{sub}=\{T_k\}\ \ with\ \ distribution\ \ T_k \sim p(T) 
\end{aligned}
\end{equation}    
    
    when each round begins, the server select a subset of tasks for task training (inner update, see algorithm \ref{algo_fl_client}), and collect all these updated parameters for meta optimization, see Algorithm \ref{algo_server}.

\begin{algorithm}
   \caption{Federated Server Aggregation}
   \label{algo_server}
   \setstretch{1.1} 
\begin{algorithmic}
   \STATE randomize initial model $G^1$
   \FOR{$each\ round\ t\gets 1,2,...,T$}
     \STATE sampel a subset of task $T_{sub}=\{T_k\}$, where $T_k \sim p(T)$
     \FOR{each task $T_i \in T$}
        \STATE $\lambda_i$, $\Delta_i$ = ClientUpdate($T_i$, $G^t$)
      \ENDFOR
      \STATE meta learner update: $G^{t+1}=G^{t}+\frac{1}{\|T_{sub}\|}\sum_{T_k \in T_{sub}}(\lambda_i*\Delta_i)$
   \ENDFOR
\end{algorithmic}
\end{algorithm}

\end{itemize}

\thispagestyle{empty}

\section{Experiments}

In this section, we present detailed experimental results to evaluate our approach. All our experiments are implemented with pytorch \cite{NIPS2019_9015}, and run on NVIDIA TeslaV100 GPU.

\subsection{Experiments set up}

\begin{figure}[h]
    \centering
    \includegraphics[width=3in]{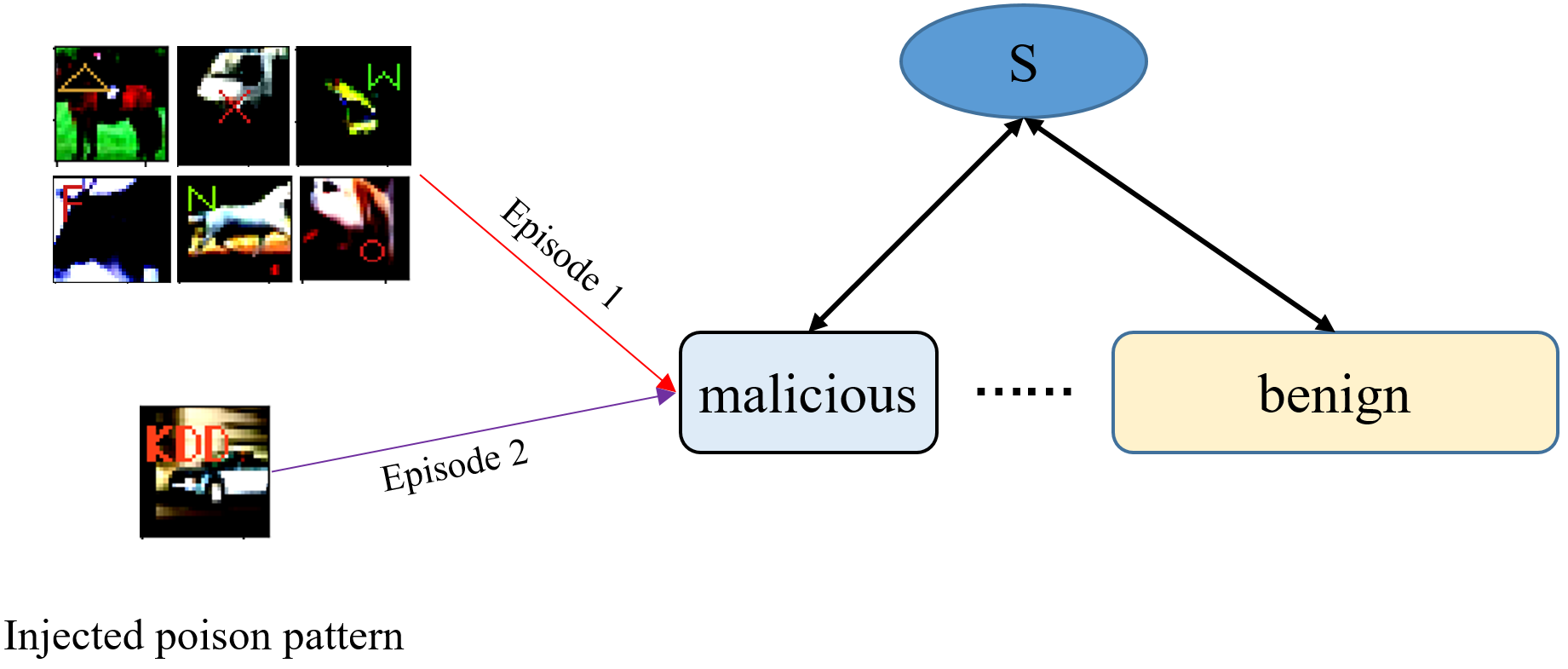}
    \caption{Poison pattern of our experiments}
    \label{fig:Poisoned_pattern}
\end{figure}

We create a federated learning cluster consists of one server and 100 clients, among of them, 6 of whom are malicious clients, the dynamic injected poison pattern for each malicious client is shown in figure \ref{fig:Poisoned_pattern}.

\begin{figure*}[htbp]
\centering
\subfigure[mnist backdoor accuracy]{
\begin{minipage}[t]{0.33\linewidth}
\centering
\includegraphics[width=2.2in, height=1.5in]{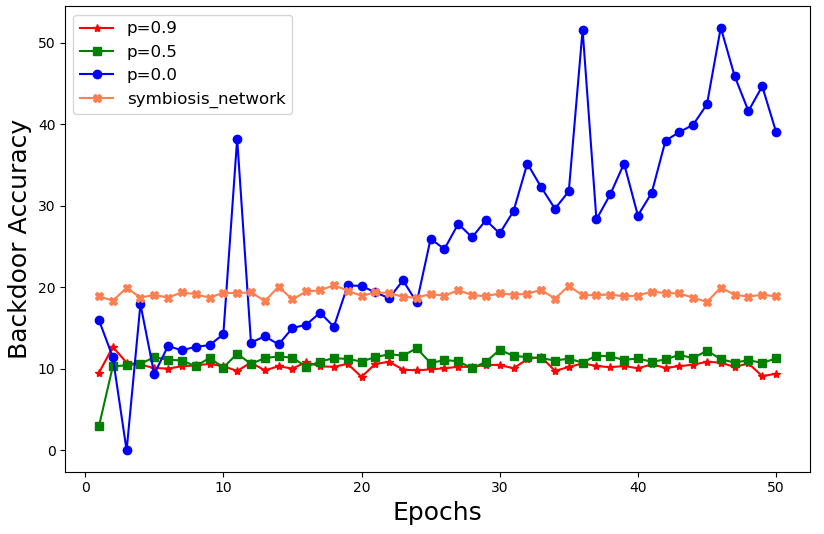}
\end{minipage}%
\label{fig:mnist_backdoor}
}%
\subfigure[cifar backdoor accuracy]{
\begin{minipage}[t]{0.33\linewidth}
\centering
\includegraphics[width=2.2in, height=1.5in]{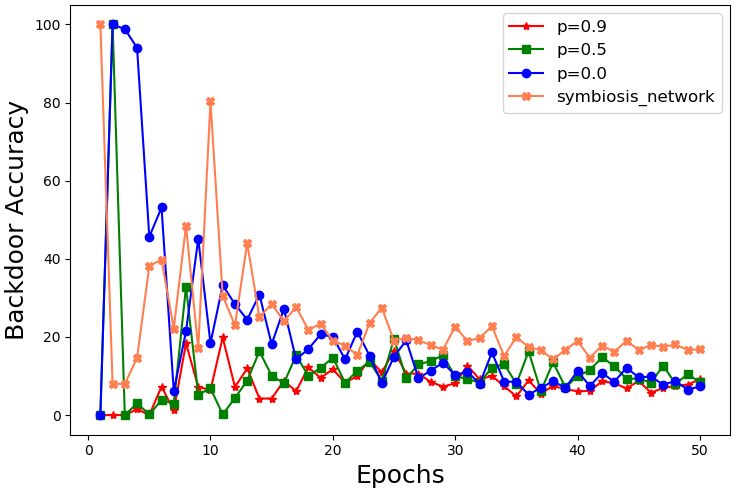}
\end{minipage}%
\label{fig:cifar_backdoor}
}%
\subfigure[tiny imagenet backdoor accuracy]{
\begin{minipage}[t]{0.33\linewidth}
\centering
\includegraphics[width=2.2in, height=1.5in]{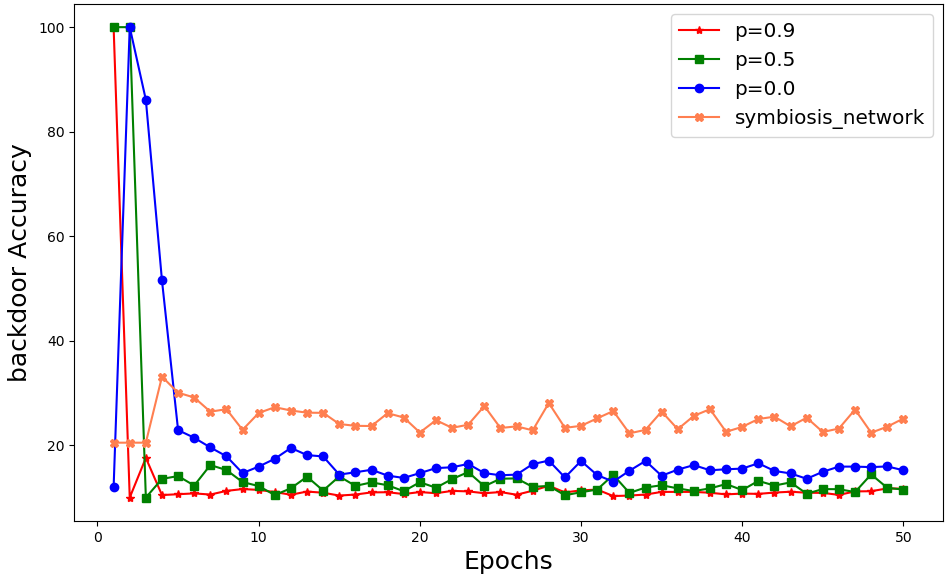}
\end{minipage}%
\label{fig:tiny_backdoor}
}%

\subfigure[mnist main accuracy]{
\begin{minipage}[t]{0.33\linewidth}
\centering
\includegraphics[width=2.2in, height=1.5in]{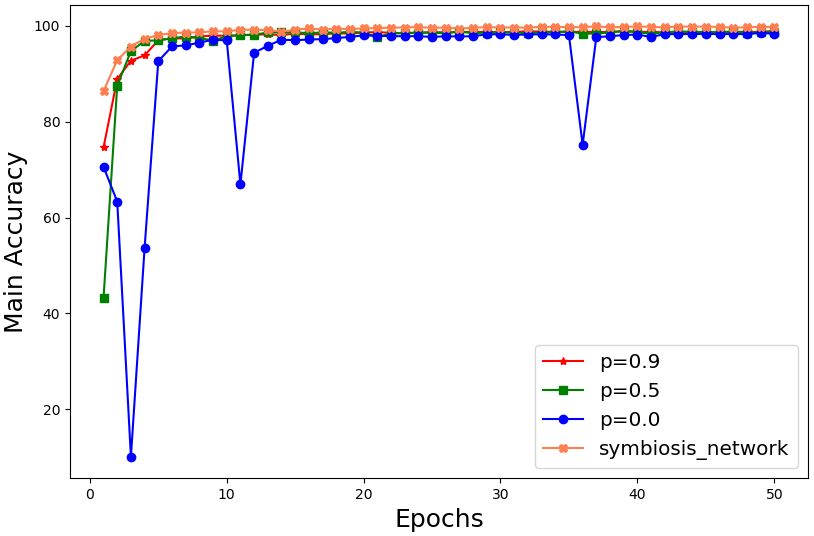}
\end{minipage}%
\label{fig:mnist_main}
}%
\subfigure[cifar main accuracy]{
\begin{minipage}[t]{0.33\linewidth}
\centering
\includegraphics[width=2.2in, height=1.5in]{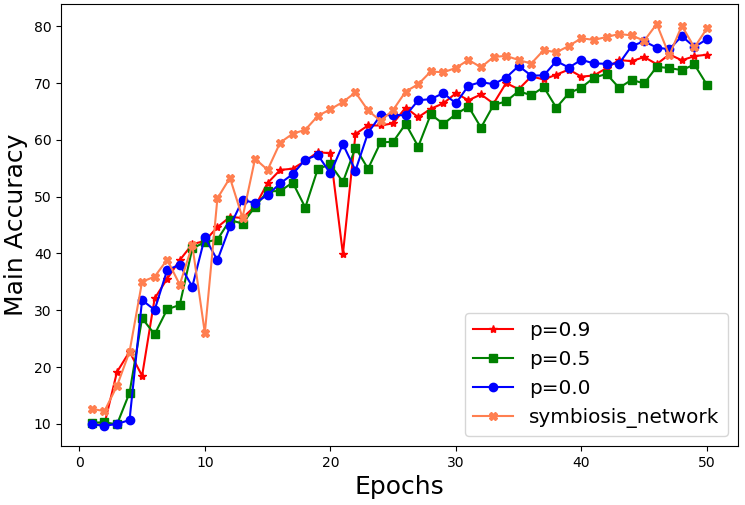}
\end{minipage}%
\label{fig:cifar_main}
}%
\subfigure[tiny imagenet main accuracy]{
\begin{minipage}[t]{0.33\linewidth}
\centering
\includegraphics[width=2.2in, height=1.5in]{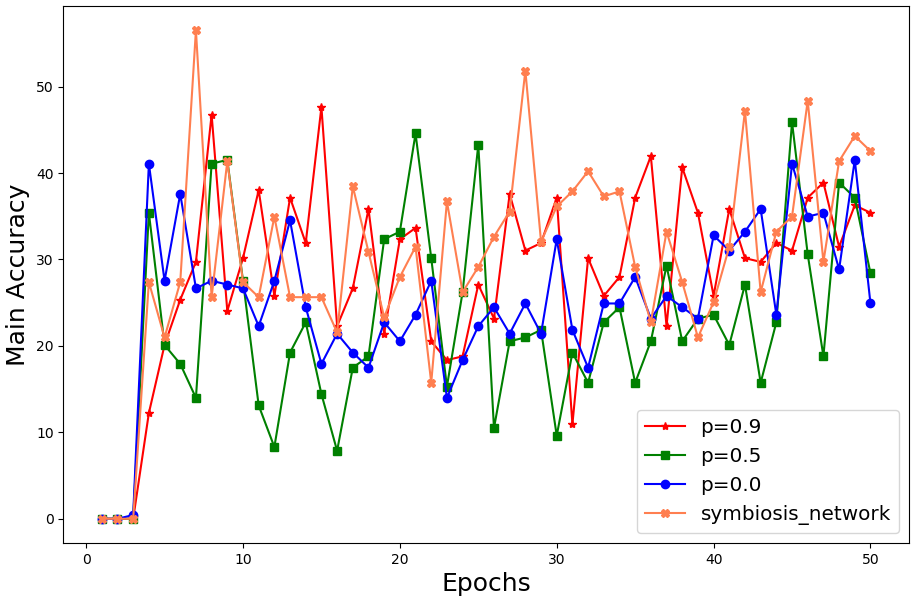}
\end{minipage}%
\label{fig:tiny_main}
}%

\centering
\caption{persist and performance evaluation} 
\label{fig:persist_and_performance}
\end{figure*}

Without loss of generality, we set $C_i\ (i=1,2,...,6)$ are malicious clients, the initial poison patterns for each malicious client are listed in table \ref{tab:initial_injected_poisons}. we split datasets using dirichlet distribution and assign them to each client respectively, for malicious clients, about 12 percent are poison datasets.

\begin{table}[h]
\renewcommand\arraystretch{1.3}
\caption{initial injected poisons}
\label{tab:initial_injected_poisons}
\centering
\begin{tabular}{lclclclclclclcl}
\hline
client & $C_1$ & $C_2$  & $C_3$ &  $C_4$ & $C_5$ & $C_6$ & \\
\hline
inject poison & $\Delta$ & $X$  & $W$ &  $F$ & $N$ & $O$ & \\
\hline
\end{tabular}

\end{table}

We choose three popular image datasets to evaluate our approach, including mnist, cifar-10 and tiny imagenet. These three datasets are increasing in size and are therefore good candidates for comparison.

\subsection{Evaluation on performance and persistent}

As shown in figure \ref{fig:persist_and_performance}, we run three different CNN architecture (LeNet for MNIST, ResNet for cifar-10 and DenseNet for tiny imagenet) to evaluate performance and persistent (see section \ref{objective}). 

Figure \ref{fig:mnist_backdoor}, \ref{fig:cifar_backdoor}, \ref{fig:tiny_backdoor} shown the backdoor accuracy performance, As previous mentioned, backdoor attack under FL setting is much harder than what we thought, the main reason is that model aggregation would cancel out most of the malicious model’s contribution and the joint model quickly forgets the backdoor, the fluctuations in the graph are due to the impact of model aggregation, we compare manually setting $p$ value \cite{DBLP:journals/corr/abs-1807-00459,Xie2020DBA:} with symbiosis network training (see equation \ref{symbiosis_network_loss}), our symbiosis network training outperform manually setting approach in most case with respect to backdoor accuracy, besides that, as the iteration progresses, this advantage can be maintained, which means that our attack approach is persistent and robust.  

Figure \ref{fig:mnist_main}, \ref{fig:cifar_main}, \ref{fig:tiny_main} shown the main task accuracy performance of our approach, as we can see, backdoor attack does not significantly affect the main task, and achieve good performances in all three datasets.

\subsection{Evaluation on fast adaptation}

We use meta optimization describe in equation \ref{meta_objective_2} as our aggregation to improve model adaptation capability, and make it quickly adapt to new poisoned task. To simulate this process, we use initial injected poisons (see table \ref{tab:initial_injected_poisons}) for malicious clients in episode 1, after that, we inject new embedded text "KDD" into local images of client $C_1$, and use it as our new poisoned datasets in episode 2.

\begin{figure*}[htbp]
\centering
\subfigure[mnist backdoor accuracy]{
\begin{minipage}[t]{0.33\linewidth}
\centering
\includegraphics[width=2.2in, height=1.5in]{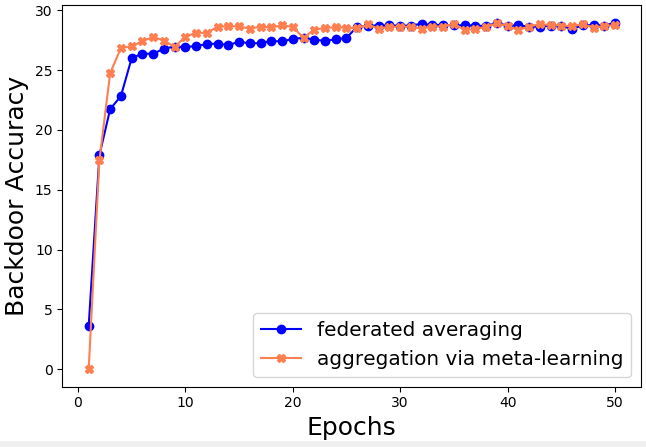}
\end{minipage}%
\label{fig:mnist_fast}
}%
\subfigure[cifar backdoor accuracy]{
\begin{minipage}[t]{0.33\linewidth}
\centering
\includegraphics[width=2.2in, height=1.5in]{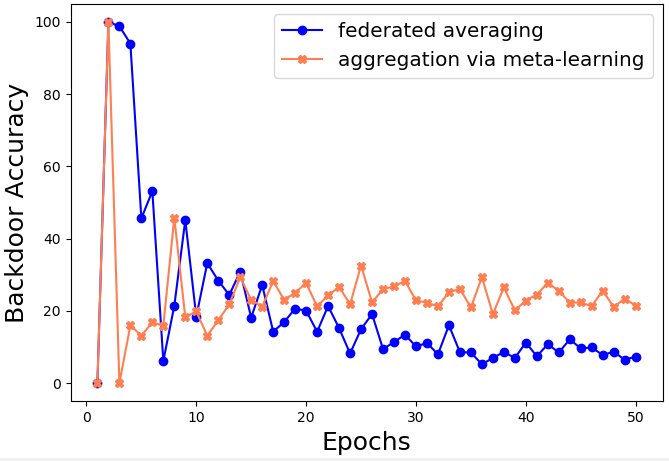}
\end{minipage}%
\label{fig:cifar_fast}
}%
\subfigure[tiny imagenet backdoor accuracy]{
\begin{minipage}[t]{0.33\linewidth}
\centering
\includegraphics[width=2.2in, height=1.5in]{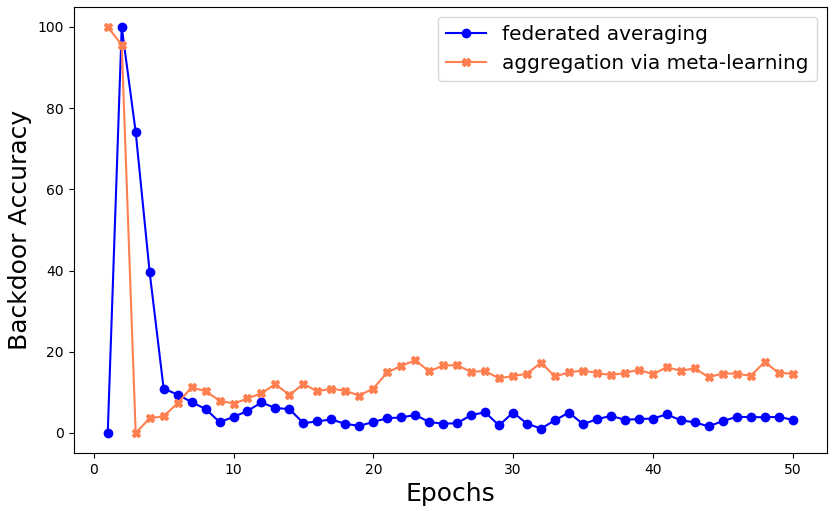}
\end{minipage}%
\label{fig:tiny_fast}
}%

\centering
\caption{fast adaptation evaluation} 
\label{fig:fast_adaptation}
\end{figure*}

Here, we use federated averaging algorithm as our baseline, the performance is shown in figure \ref{fig:fast_adaptation}, after the first few rounds, the meta-learning method quickly surpassed the federated averaging and achieve the same results with fewer steps.

\thispagestyle{empty}

\section{Conclusion and future works}

Federated learning is appealing because of its confidentiality and scalability, although adversarial attacks under federated learning setting has been studied extensively, it is still mainly focus on static scenarios. Dynamic backdoor attacks, on the other hand, are more challenging and ubiquitous in our real world.   

In this paper, we introduce dynamic backdoor attacks problem under federated learning setting, and propose three corresponding objectives, coupled with detailed definitions and solutions for each of them, finally, we give an efficient and feasible solution to solve this problem. In future work, We intend to improve our work from the following two aspects:

\begin{itemize}
    \item Our experiments mainly focus on image classification problems, we will verify the correctness of our algorithm with more experimental results.
    
    \item Explore how to improve other aggregation algorithms so that it can be compatible with meta-learning framework.
\end{itemize}

\bibliographystyle{ACM-Reference-Format}
\bibliography{sample-base}


\begin{thebibliography}{31}


\ifx \showCODEN    \undefined \def \showCODEN     #1{\unskip}     \fi
\ifx \showDOI      \undefined \def \showDOI       #1{#1}\fi
\ifx \showISBNx    \undefined \def \showISBNx     #1{\unskip}     \fi
\ifx \showISBNxiii \undefined \def \showISBNxiii  #1{\unskip}     \fi
\ifx \showISSN     \undefined \def \showISSN      #1{\unskip}     \fi
\ifx \showLCCN     \undefined \def \showLCCN      #1{\unskip}     \fi
\ifx \shownote     \undefined \def \shownote      #1{#1}          \fi
\ifx \showarticletitle \undefined \def \showarticletitle #1{#1}   \fi
\ifx \showURL      \undefined \def \showURL       {\relax}        \fi
\providecommand\bibfield[2]{#2}
\providecommand\bibinfo[2]{#2}
\providecommand\natexlab[1]{#1}
\providecommand\showeprint[2][]{arXiv:#2}

\bibitem[\protect\citeauthoryear{Bagdasaryan, Veit, Hua, Estrin, and
  Shmatikov}{Bagdasaryan et~al\mbox{.}}{2018}]%
        {DBLP:journals/corr/abs-1807-00459}
\bibfield{author}{\bibinfo{person}{Eugene Bagdasaryan},
  \bibinfo{person}{Andreas Veit}, \bibinfo{person}{Yiqing Hua},
  \bibinfo{person}{Deborah Estrin}, {and} \bibinfo{person}{Vitaly Shmatikov}.}
  \bibinfo{year}{2018}\natexlab{}.
\newblock \showarticletitle{How To Backdoor Federated Learning}.
\newblock \bibinfo{journal}{\emph{CoRR}}  \bibinfo{volume}{abs/1807.00459}
  (\bibinfo{year}{2018}).
\newblock
\showeprint[arxiv]{1807.00459}


\bibitem[\protect\citeauthoryear{Bhagoji, Chakraborty, Mittal, and
  Calo}{Bhagoji et~al\mbox{.}}{2018}]%
        {DBLP:journals/corr/abs-1811-12470}
\bibfield{author}{\bibinfo{person}{Arjun~Nitin Bhagoji},
  \bibinfo{person}{Supriyo Chakraborty}, \bibinfo{person}{Prateek Mittal},
  {and} \bibinfo{person}{Seraphin~B. Calo}.} \bibinfo{year}{2018}\natexlab{}.
\newblock \showarticletitle{Analyzing Federated Learning through an Adversarial
  Lens}.
\newblock \bibinfo{journal}{\emph{CoRR}}  \bibinfo{volume}{abs/1811.12470}
  (\bibinfo{year}{2018}).
\newblock
\showeprint[arxiv]{1811.12470}


\bibitem[\protect\citeauthoryear{Bonawitz, Eichner, Grieskamp, Huba, Ingerman,
  Ivanov, Kiddon, Konecn{\'{y}}, Mazzocchi, McMahan, Overveldt, Petrou, Ramage,
  and Roselander}{Bonawitz et~al\mbox{.}}{2019}]%
        {DBLP:journals/corr/abs-1902-01046}
\bibfield{author}{\bibinfo{person}{Keith Bonawitz}, \bibinfo{person}{Hubert
  Eichner}, \bibinfo{person}{Wolfgang Grieskamp}, \bibinfo{person}{Dzmitry
  Huba}, \bibinfo{person}{Alex Ingerman}, \bibinfo{person}{Vladimir Ivanov},
  \bibinfo{person}{Chlo{\'{e}} Kiddon}, \bibinfo{person}{Jakub Konecn{\'{y}}},
  \bibinfo{person}{Stefano Mazzocchi}, \bibinfo{person}{H.~Brendan McMahan},
  \bibinfo{person}{Timon~Van Overveldt}, \bibinfo{person}{David Petrou},
  \bibinfo{person}{Daniel Ramage}, {and} \bibinfo{person}{Jason Roselander}.}
  \bibinfo{year}{2019}\natexlab{}.
\newblock \showarticletitle{Towards Federated Learning at Scale: System
  Design}.
\newblock \bibinfo{journal}{\emph{CoRR}}  \bibinfo{volume}{abs/1902.01046}
  (\bibinfo{year}{2019}).
\newblock
\showeprint[arxiv]{1902.01046}


\bibitem[\protect\citeauthoryear{Chen, Dong, Li, and He}{Chen
  et~al\mbox{.}}{2018}]%
        {DBLP:journals/corr/abs-1802-07876}
\bibfield{author}{\bibinfo{person}{Fei Chen}, \bibinfo{person}{Zhenhua Dong},
  \bibinfo{person}{Zhenguo Li}, {and} \bibinfo{person}{Xiuqiang He}.}
  \bibinfo{year}{2018}\natexlab{}.
\newblock \showarticletitle{Federated Meta-Learning for Recommendation}.
\newblock \bibinfo{journal}{\emph{CoRR}}  \bibinfo{volume}{abs/1802.07876}
  (\bibinfo{year}{2018}).
\newblock
\showeprint[arxiv]{1802.07876}
\urldef\tempurl%
\url{http://arxiv.org/abs/1802.07876}
\showURL{%
\tempurl}


\bibitem[\protect\citeauthoryear{Devlin, Chang, Lee, and Toutanova}{Devlin
  et~al\mbox{.}}{2018}]%
        {DBLP:journals/corr/abs-1810-04805}
\bibfield{author}{\bibinfo{person}{Jacob Devlin}, \bibinfo{person}{Ming{-}Wei
  Chang}, \bibinfo{person}{Kenton Lee}, {and} \bibinfo{person}{Kristina
  Toutanova}.} \bibinfo{year}{2018}\natexlab{}.
\newblock \showarticletitle{{BERT:} Pre-training of Deep Bidirectional
  Transformers for Language Understanding}.
\newblock \bibinfo{journal}{\emph{CoRR}}  \bibinfo{volume}{abs/1810.04805}
  (\bibinfo{year}{2018}).
\newblock
\showeprint[arxiv]{1810.04805}


\bibitem[\protect\citeauthoryear{Finn, Abbeel, and Levine}{Finn
  et~al\mbox{.}}{2017}]%
        {DBLP:journals/corr/FinnAL17}
\bibfield{author}{\bibinfo{person}{Chelsea Finn}, \bibinfo{person}{Pieter
  Abbeel}, {and} \bibinfo{person}{Sergey Levine}.}
  \bibinfo{year}{2017}\natexlab{}.
\newblock \showarticletitle{Model-Agnostic Meta-Learning for Fast Adaptation of
  Deep Networks}.
\newblock \bibinfo{journal}{\emph{CoRR}}  \bibinfo{volume}{abs/1703.03400}
  (\bibinfo{year}{2017}).
\newblock
\showeprint[arxiv]{1703.03400}
\urldef\tempurl%
\url{http://arxiv.org/abs/1703.03400}
\showURL{%
\tempurl}


\bibitem[\protect\citeauthoryear{Graves, Wayne, and Danihelka}{Graves
  et~al\mbox{.}}{2014}]%
        {graves2014neural}
\bibfield{author}{\bibinfo{person}{Alex Graves}, \bibinfo{person}{Greg Wayne},
  {and} \bibinfo{person}{Ivo Danihelka}.} \bibinfo{year}{2014}\natexlab{}.
\newblock \bibinfo{title}{Neural Turing Machines}.
\newblock
\newblock
\urldef\tempurl%
\url{http://arxiv.org/abs/1410.5401}
\showURL{%
\tempurl}
\newblock
\shownote{cite arxiv:1410.5401.}


\bibitem[\protect\citeauthoryear{Hannun, Case, Casper, Catanzaro, Diamos,
  Elsen, Prenger, Satheesh, Sengupta, Coates, and Ng}{Hannun
  et~al\mbox{.}}{2014}]%
        {DBLP:journals/corr/HannunCCCDEPSSCN14}
\bibfield{author}{\bibinfo{person}{Awni~Y. Hannun}, \bibinfo{person}{Carl
  Case}, \bibinfo{person}{Jared Casper}, \bibinfo{person}{Bryan Catanzaro},
  \bibinfo{person}{Greg Diamos}, \bibinfo{person}{Erich Elsen},
  \bibinfo{person}{Ryan Prenger}, \bibinfo{person}{Sanjeev Satheesh},
  \bibinfo{person}{Shubho Sengupta}, \bibinfo{person}{Adam Coates}, {and}
  \bibinfo{person}{Andrew~Y. Ng}.} \bibinfo{year}{2014}\natexlab{}.
\newblock \showarticletitle{Deep Speech: Scaling up end-to-end speech
  recognition}.
\newblock \bibinfo{journal}{\emph{CoRR}}  \bibinfo{volume}{abs/1412.5567}
  (\bibinfo{year}{2014}).
\newblock
\showeprint[arxiv]{1412.5567}


\bibitem[\protect\citeauthoryear{He, Zhang, Ren, and Sun}{He
  et~al\mbox{.}}{2015}]%
        {DBLP:journals/corr/HeZRS15}
\bibfield{author}{\bibinfo{person}{Kaiming He}, \bibinfo{person}{Xiangyu
  Zhang}, \bibinfo{person}{Shaoqing Ren}, {and} \bibinfo{person}{Jian Sun}.}
  \bibinfo{year}{2015}\natexlab{}.
\newblock \showarticletitle{Deep Residual Learning for Image Recognition}.
\newblock \bibinfo{journal}{\emph{CoRR}}  \bibinfo{volume}{abs/1512.03385}
  (\bibinfo{year}{2015}).
\newblock
\showeprint[arxiv]{1512.03385}


\bibitem[\protect\citeauthoryear{Kairouz, McMahan, Avent, Bellet, Bennis,
  Bhagoji, Bonawitz, Charles, Cormode, Cummings, et~al\mbox{.}}{Kairouz
  et~al\mbox{.}}{2019}]%
        {kairouz2019advances}
\bibfield{author}{\bibinfo{person}{Peter Kairouz}, \bibinfo{person}{H~Brendan
  McMahan}, \bibinfo{person}{Brendan Avent}, \bibinfo{person}{Aur{\'e}lien
  Bellet}, \bibinfo{person}{Mehdi Bennis}, \bibinfo{person}{Arjun~Nitin
  Bhagoji}, \bibinfo{person}{Keith Bonawitz}, \bibinfo{person}{Zachary
  Charles}, \bibinfo{person}{Graham Cormode}, \bibinfo{person}{Rachel
  Cummings}, {et~al\mbox{.}}} \bibinfo{year}{2019}\natexlab{}.
\newblock \showarticletitle{Advances and open problems in federated learning}.
\newblock \bibinfo{journal}{\emph{arXiv preprint arXiv:1912.04977}}
  (\bibinfo{year}{2019}).
\newblock


\bibitem[\protect\citeauthoryear{Koch, Zemel, and Salakhutdinov}{Koch
  et~al\mbox{.}}{2015}]%
        {Koch2015SiameseNN}
\bibfield{author}{\bibinfo{person}{Gregory Koch}, \bibinfo{person}{Richard
  Zemel}, {and} \bibinfo{person}{Ruslan Salakhutdinov}.}
  \bibinfo{year}{2015}\natexlab{}.
\newblock \showarticletitle{Siamese Neural Networks for One-shot Image
  Recognition}.
\newblock


\bibitem[\protect\citeauthoryear{Krizhevsky, Sutskever, and Hinton}{Krizhevsky
  et~al\mbox{.}}{2012}]%
        {Krizhevsky:2012:ICD:2999134.2999257}
\bibfield{author}{\bibinfo{person}{Alex Krizhevsky}, \bibinfo{person}{Ilya
  Sutskever}, {and} \bibinfo{person}{Geoffrey~E. Hinton}.}
  \bibinfo{year}{2012}\natexlab{}.
\newblock \showarticletitle{ImageNet Classification with Deep Convolutional
  Neural Networks}. In \bibinfo{booktitle}{\emph{Proceedings of the 25th
  International Conference on Neural Information Processing Systems - Volume
  1}} (Lake Tahoe, Nevada) \emph{(\bibinfo{series}{NIPS'12})}.
  \bibinfo{publisher}{Curran Associates Inc.}, \bibinfo{address}{USA},
  \bibinfo{pages}{1097--1105}.
\newblock


\bibitem[\protect\citeauthoryear{LeCun, Bengio, and Hinton}{LeCun
  et~al\mbox{.}}{2015}]%
        {lecun2015deeplearning}
\bibfield{author}{\bibinfo{person}{Yann LeCun}, \bibinfo{person}{Yoshua
  Bengio}, {and} \bibinfo{person}{Geoffrey Hinton}.}
  \bibinfo{year}{2015}\natexlab{}.
\newblock \showarticletitle{Deep Learning}.
\newblock \bibinfo{journal}{\emph{Nature}} \bibinfo{volume}{521},
  \bibinfo{number}{7553} (\bibinfo{year}{2015}), \bibinfo{pages}{436--444}.
\newblock
\urldef\tempurl%
\url{https://doi.org/10.1038/nature14539}
\showDOI{\tempurl}


\bibitem[\protect\citeauthoryear{Li, Milletar{\`\i}, Xu, Rieke, Hancox, Zhu,
  Baust, Cheng, Ourselin, Cardoso, et~al\mbox{.}}{Li et~al\mbox{.}}{2019}]%
        {li2019privacy}
\bibfield{author}{\bibinfo{person}{Wenqi Li}, \bibinfo{person}{Fausto
  Milletar{\`\i}}, \bibinfo{person}{Daguang Xu}, \bibinfo{person}{Nicola
  Rieke}, \bibinfo{person}{Jonny Hancox}, \bibinfo{person}{Wentao Zhu},
  \bibinfo{person}{Maximilian Baust}, \bibinfo{person}{Yan Cheng},
  \bibinfo{person}{S{\'e}bastien Ourselin}, \bibinfo{person}{M~Jorge Cardoso},
  {et~al\mbox{.}}} \bibinfo{year}{2019}\natexlab{}.
\newblock \showarticletitle{Privacy-preserving Federated Brain Tumour
  Segmentation}. In \bibinfo{booktitle}{\emph{International Workshop on Machine
  Learning in Medical Imaging}}. Springer, \bibinfo{pages}{133--141}.
\newblock


\bibitem[\protect\citeauthoryear{Liu, Huang, Luo, Huang, Liu, Chen, Feng, Chen,
  Yu, and Yang}{Liu et~al\mbox{.}}{2020}]%
        {liu2020fedvision}
\bibfield{author}{\bibinfo{person}{Yang Liu}, \bibinfo{person}{Anbu Huang},
  \bibinfo{person}{Yun Luo}, \bibinfo{person}{He Huang},
  \bibinfo{person}{Youzhi Liu}, \bibinfo{person}{Yuanyuan Chen},
  \bibinfo{person}{Lican Feng}, \bibinfo{person}{Tianjian Chen},
  \bibinfo{person}{Han Yu}, {and} \bibinfo{person}{Qiang Yang}.}
  \bibinfo{year}{2020}\natexlab{}.
\newblock \showarticletitle{FedVision: An Online Visual Object Detection
  Platform Powered by Federated Learning}.
\newblock \bibinfo{journal}{\emph{arXiv preprint arXiv:2001.06202}}
  (\bibinfo{year}{2020}).
\newblock


\bibitem[\protect\citeauthoryear{McMahan, Moore, Ramage, Hampson, and
  y~Arcas}{McMahan et~al\mbox{.}}{2017}]%
        {pmlr-v54-mcmahan17a}
\bibfield{author}{\bibinfo{person}{Brendan McMahan}, \bibinfo{person}{Eider
  Moore}, \bibinfo{person}{Daniel Ramage}, \bibinfo{person}{Seth Hampson},
  {and} \bibinfo{person}{Blaise~Aguera y Arcas}.}
  \bibinfo{year}{2017}\natexlab{}.
\newblock \showarticletitle{{Communication-Efficient Learning of Deep Networks
  from Decentralized Data}}. In \bibinfo{booktitle}{\emph{Proceedings of the
  20th International Conference on Artificial Intelligence and Statistics}}
  \emph{(\bibinfo{series}{Proceedings of Machine Learning Research})},
  \bibfield{editor}{\bibinfo{person}{Aarti Singh} {and} \bibinfo{person}{Jerry
  Zhu}} (Eds.), Vol.~\bibinfo{volume}{54}. \bibinfo{publisher}{PMLR},
  \bibinfo{address}{Fort Lauderdale, FL, USA}, \bibinfo{pages}{1273--1282}.
\newblock


\bibitem[\protect\citeauthoryear{Mikolov, Chen, Corrado, and Dean}{Mikolov
  et~al\mbox{.}}{2013}]%
        {mikolov2013efficient}
\bibfield{author}{\bibinfo{person}{Tomas Mikolov}, \bibinfo{person}{Kai Chen},
  \bibinfo{person}{Greg Corrado}, {and} \bibinfo{person}{Jeffrey Dean}.}
  \bibinfo{year}{2013}\natexlab{}.
\newblock \bibinfo{title}{Efficient Estimation of Word Representations in
  Vector Space}.
\newblock
\newblock
\newblock
\shownote{cite arxiv:1301.3781.}


\bibitem[\protect\citeauthoryear{Nichol, Achiam, and Schulman}{Nichol
  et~al\mbox{.}}{2018}]%
        {DBLP:journals/corr/abs-1803-02999}
\bibfield{author}{\bibinfo{person}{Alex Nichol}, \bibinfo{person}{Joshua
  Achiam}, {and} \bibinfo{person}{John Schulman}.}
  \bibinfo{year}{2018}\natexlab{}.
\newblock \showarticletitle{On First-Order Meta-Learning Algorithms}.
\newblock \bibinfo{journal}{\emph{CoRR}}  \bibinfo{volume}{abs/1803.02999}
  (\bibinfo{year}{2018}).
\newblock
\showeprint[arxiv]{1803.02999}
\urldef\tempurl%
\url{http://arxiv.org/abs/1803.02999}
\showURL{%
\tempurl}


\bibitem[\protect\citeauthoryear{Paszke, Gross, Massa, Lerer, Bradbury, Chanan,
  Killeen, Lin, Gimelshein, Antiga, Desmaison, Kopf, Yang, DeVito, Raison,
  Tejani, Chilamkurthy, Steiner, Fang, Bai, and Chintala}{Paszke
  et~al\mbox{.}}{2019}]%
        {NIPS2019_9015}
\bibfield{author}{\bibinfo{person}{Adam Paszke}, \bibinfo{person}{Sam Gross},
  \bibinfo{person}{Francisco Massa}, \bibinfo{person}{Adam Lerer},
  \bibinfo{person}{James Bradbury}, \bibinfo{person}{Gregory Chanan},
  \bibinfo{person}{Trevor Killeen}, \bibinfo{person}{Zeming Lin},
  \bibinfo{person}{Natalia Gimelshein}, \bibinfo{person}{Luca Antiga},
  \bibinfo{person}{Alban Desmaison}, \bibinfo{person}{Andreas Kopf},
  \bibinfo{person}{Edward Yang}, \bibinfo{person}{Zachary DeVito},
  \bibinfo{person}{Martin Raison}, \bibinfo{person}{Alykhan Tejani},
  \bibinfo{person}{Sasank Chilamkurthy}, \bibinfo{person}{Benoit Steiner},
  \bibinfo{person}{Lu Fang}, \bibinfo{person}{Junjie Bai}, {and}
  \bibinfo{person}{Soumith Chintala}.} \bibinfo{year}{2019}\natexlab{}.
\newblock \showarticletitle{PyTorch: An Imperative Style, High-Performance Deep
  Learning Library}.
\newblock In \bibinfo{booktitle}{\emph{Advances in Neural Information
  Processing Systems 32}}, \bibfield{editor}{\bibinfo{person}{H.~Wallach},
  \bibinfo{person}{H.~Larochelle}, \bibinfo{person}{A.~Beygelzimer},
  \bibinfo{person}{F.~d\textquotesingle Alch\'{e}-Buc},
  \bibinfo{person}{E.~Fox}, {and} \bibinfo{person}{R.~Garnett}} (Eds.).
  \bibinfo{publisher}{Curran Associates, Inc.}, \bibinfo{pages}{8026--8037}.
\newblock
\urldef\tempurl%
\url{http://papers.nips.cc/paper/9015-pytorch-an-imperative-style-high-performance-deep-learning-library.pdf}
\showURL{%
\tempurl}


\bibitem[\protect\citeauthoryear{Redmon and Farhadi}{Redmon and
  Farhadi}{2018}]%
        {redmon2018yolov3}
\bibfield{author}{\bibinfo{person}{Joseph Redmon} {and} \bibinfo{person}{Ali
  Farhadi}.} \bibinfo{year}{2018}\natexlab{}.
\newblock \showarticletitle{Yolov3: An incremental improvement}.
\newblock \bibinfo{journal}{\emph{arXiv preprint arXiv:1804.02767}}
  (\bibinfo{year}{2018}).
\newblock


\bibitem[\protect\citeauthoryear{Snell, Swersky, and Zemel}{Snell
  et~al\mbox{.}}{2017}]%
        {DBLP:journals/corr/SnellSZ17}
\bibfield{author}{\bibinfo{person}{Jake Snell}, \bibinfo{person}{Kevin
  Swersky}, {and} \bibinfo{person}{Richard~S. Zemel}.}
  \bibinfo{year}{2017}\natexlab{}.
\newblock \showarticletitle{Prototypical Networks for Few-shot Learning}.
\newblock \bibinfo{journal}{\emph{CoRR}}  \bibinfo{volume}{abs/1703.05175}
  (\bibinfo{year}{2017}).
\newblock
\showeprint[arxiv]{1703.05175}
\urldef\tempurl%
\url{http://arxiv.org/abs/1703.05175}
\showURL{%
\tempurl}


\bibitem[\protect\citeauthoryear{Sun, Kairouz, Suresh, and McMahan}{Sun
  et~al\mbox{.}}{2019}]%
        {sun2019can}
\bibfield{author}{\bibinfo{person}{Ziteng Sun}, \bibinfo{person}{Peter
  Kairouz}, \bibinfo{person}{Ananda~Theertha Suresh}, {and}
  \bibinfo{person}{H~Brendan McMahan}.} \bibinfo{year}{2019}\natexlab{}.
\newblock \showarticletitle{Can You Really Backdoor Federated Learning?}
\newblock \bibinfo{journal}{\emph{arXiv preprint arXiv:1911.07963}}
  (\bibinfo{year}{2019}).
\newblock


\bibitem[\protect\citeauthoryear{Sung, Yang, Zhang, Xiang, Torr, and
  Hospedales}{Sung et~al\mbox{.}}{2017}]%
        {DBLP:journals/corr/abs-1711-06025}
\bibfield{author}{\bibinfo{person}{Flood Sung}, \bibinfo{person}{Yongxin Yang},
  \bibinfo{person}{Li Zhang}, \bibinfo{person}{Tao Xiang},
  \bibinfo{person}{Philip H.~S. Torr}, {and} \bibinfo{person}{Timothy~M.
  Hospedales}.} \bibinfo{year}{2017}\natexlab{}.
\newblock \showarticletitle{Learning to Compare: Relation Network for Few-Shot
  Learning}.
\newblock \bibinfo{journal}{\emph{CoRR}}  \bibinfo{volume}{abs/1711.06025}
  (\bibinfo{year}{2017}).
\newblock
\showeprint[arxiv]{1711.06025}
\urldef\tempurl%
\url{http://arxiv.org/abs/1711.06025}
\showURL{%
\tempurl}


\bibitem[\protect\citeauthoryear{van~den Oord, Dieleman, Zen, Simonyan,
  Vinyals, Graves, Kalchbrenner, Senior, and Kavukcuoglu}{van~den Oord
  et~al\mbox{.}}{2016}]%
        {DBLP:journals/corr/OordDZSVGKSK16}
\bibfield{author}{\bibinfo{person}{A{\"{a}}ron van~den Oord},
  \bibinfo{person}{Sander Dieleman}, \bibinfo{person}{Heiga Zen},
  \bibinfo{person}{Karen Simonyan}, \bibinfo{person}{Oriol Vinyals},
  \bibinfo{person}{Alex Graves}, \bibinfo{person}{Nal Kalchbrenner},
  \bibinfo{person}{Andrew~W. Senior}, {and} \bibinfo{person}{Koray
  Kavukcuoglu}.} \bibinfo{year}{2016}\natexlab{}.
\newblock \showarticletitle{WaveNet: {A} Generative Model for Raw Audio}.
\newblock \bibinfo{journal}{\emph{CoRR}}  \bibinfo{volume}{abs/1609.03499}
  (\bibinfo{year}{2016}).
\newblock
\showeprint[arxiv]{1609.03499}


\bibitem[\protect\citeauthoryear{Vaswani, Shazeer, Parmar, Uszkoreit, Jones,
  Gomez, Kaiser, and Polosukhin}{Vaswani et~al\mbox{.}}{2017}]%
        {NIPS2017_7181}
\bibfield{author}{\bibinfo{person}{Ashish Vaswani}, \bibinfo{person}{Noam
  Shazeer}, \bibinfo{person}{Niki Parmar}, \bibinfo{person}{Jakob Uszkoreit},
  \bibinfo{person}{Llion Jones}, \bibinfo{person}{Aidan~N Gomez},
  \bibinfo{person}{\L~ukasz Kaiser}, {and} \bibinfo{person}{Illia Polosukhin}.}
  \bibinfo{year}{2017}\natexlab{}.
\newblock \showarticletitle{Attention is All you Need}.
\newblock In \bibinfo{booktitle}{\emph{Advances in Neural Information
  Processing Systems 30}}, \bibfield{editor}{\bibinfo{person}{I.~Guyon},
  \bibinfo{person}{U.~V. Luxburg}, \bibinfo{person}{S.~Bengio},
  \bibinfo{person}{H.~Wallach}, \bibinfo{person}{R.~Fergus},
  \bibinfo{person}{S.~Vishwanathan}, {and} \bibinfo{person}{R.~Garnett}}
  (Eds.). \bibinfo{publisher}{Curran Associates, Inc.},
  \bibinfo{pages}{5998--6008}.
\newblock
\urldef\tempurl%
\url{http://papers.nips.cc/paper/7181-attention-is-all-you-need.pdf}
\showURL{%
\tempurl}


\bibitem[\protect\citeauthoryear{Vinyals, Blundell, Lillicrap, Kavukcuoglu, and
  Wierstra}{Vinyals et~al\mbox{.}}{2016}]%
        {DBLP:journals/corr/VinyalsBLKW16}
\bibfield{author}{\bibinfo{person}{Oriol Vinyals}, \bibinfo{person}{Charles
  Blundell}, \bibinfo{person}{Timothy~P. Lillicrap}, \bibinfo{person}{Koray
  Kavukcuoglu}, {and} \bibinfo{person}{Daan Wierstra}.}
  \bibinfo{year}{2016}\natexlab{}.
\newblock \showarticletitle{Matching Networks for One Shot Learning}.
\newblock \bibinfo{journal}{\emph{CoRR}}  \bibinfo{volume}{abs/1606.04080}
  (\bibinfo{year}{2016}).
\newblock
\showeprint[arxiv]{1606.04080}
\urldef\tempurl%
\url{http://arxiv.org/abs/1606.04080}
\showURL{%
\tempurl}


\bibitem[\protect\citeauthoryear{Voigt and Bussche}{Voigt and Bussche}{2017}]%
        {Voigt:2017:EGD:3152676}
\bibfield{author}{\bibinfo{person}{Paul Voigt} {and} \bibinfo{person}{Axel
  von~dem Bussche}.} \bibinfo{year}{2017}\natexlab{}.
\newblock \bibinfo{booktitle}{\emph{The EU General Data Protection Regulation
  (GDPR): A Practical Guide} (\bibinfo{edition}{1st} ed.)}.
\newblock \bibinfo{publisher}{Springer Publishing Company, Incorporated}.
\newblock
\showISBNx{3319579584, 9783319579580}


\bibitem[\protect\citeauthoryear{Weston, Chopra, and Bordes}{Weston
  et~al\mbox{.}}{2014}]%
        {weston2014memory}
\bibfield{author}{\bibinfo{person}{Jason Weston}, \bibinfo{person}{Sumit
  Chopra}, {and} \bibinfo{person}{Antoine Bordes}.}
  \bibinfo{year}{2014}\natexlab{}.
\newblock \showarticletitle{Memory Networks}.
\newblock  (\bibinfo{year}{2014}).
\newblock
\urldef\tempurl%
\url{http://arxiv.org/abs/1410.3916}
\showURL{%
\tempurl}
\newblock
\shownote{cite arxiv:1410.3916.}


\bibitem[\protect\citeauthoryear{Xie, Huang, Chen, and Li}{Xie
  et~al\mbox{.}}{2020}]%
        {Xie2020DBA:}
\bibfield{author}{\bibinfo{person}{Chulin Xie}, \bibinfo{person}{Keli Huang},
  \bibinfo{person}{Pin-Yu Chen}, {and} \bibinfo{person}{Bo Li}.}
  \bibinfo{year}{2020}\natexlab{}.
\newblock \showarticletitle{DBA: Distributed Backdoor Attacks against Federated
  Learning}. In \bibinfo{booktitle}{\emph{International Conference on Learning
  Representations}}.
\newblock
\urldef\tempurl%
\url{https://openreview.net/forum?id=rkgyS0VFvr}
\showURL{%
\tempurl}


\bibitem[\protect\citeauthoryear{Yang, Liu, Chen, and Tong}{Yang
  et~al\mbox{.}}{2019}]%
        {DBLP:journals/corr/abs-1902-04885}
\bibfield{author}{\bibinfo{person}{Qiang Yang}, \bibinfo{person}{Yang Liu},
  \bibinfo{person}{Tianjian Chen}, {and} \bibinfo{person}{Yongxin Tong}.}
  \bibinfo{year}{2019}\natexlab{}.
\newblock \showarticletitle{Federated Machine Learning: Concept and
  Applications}.
\newblock \bibinfo{journal}{\emph{CoRR}}  \bibinfo{volume}{abs/1902.04885}
  (\bibinfo{year}{2019}).
\newblock
\showeprint[arxiv]{1902.04885}


\bibitem[\protect\citeauthoryear{Zintgraf, Shiarlis, Kurin, Hofmann, and
  Whiteson}{Zintgraf et~al\mbox{.}}{2019}]%
        {zintgraf2019fast}
\bibfield{author}{\bibinfo{person}{Luisa Zintgraf}, \bibinfo{person}{Kyriacos
  Shiarlis}, \bibinfo{person}{Vitaly Kurin}, \bibinfo{person}{Katja Hofmann},
  {and} \bibinfo{person}{Shimon Whiteson}.} \bibinfo{year}{2019}\natexlab{}.
\newblock \showarticletitle{Fast Context Adaptation via Meta-Learning}. In
  \bibinfo{booktitle}{\emph{Thirty-sixth International Conference on Machine
  Learning (ICML)}}.
\newblock


\end{thebibliography}

\appendix

\thispagestyle{empty}

\end{document}